\documentclass[letterpaper, 10 pt, conference]{ieeeconf}  
\IEEEoverridecommandlockouts                              
\usepackage{graphics}    
\usepackage{times}       
\usepackage{amsmath}     
\usepackage{amssymb}     
\usepackage{graphicx}
\usepackage{xcolor}
\usepackage{algorithm}
\usepackage[noend]{algpseudocode}
\usepackage{url}
\usepackage{adjustbox} 

\usepackage[font=small]{caption}

\def\figref#1{Fig.~\ref{#1}}

\def\eqref#1{Eq.~(\ref{#1})}

\newcommand\etal{\emph{et al.}}

\title{\LARGE \bf Efficient Trajectory Optimization for Autonomous Racing via Formula-1 Data-Driven Initialization
}
\author{Anonymous Authors
\vspace{-0.5cm}
}
\author{Samir Shehadeh \and Lukas Kutsch \and Nils Dengler \and Sicong Pan \and Maren Bennewitz
  \thanks{All authors are with the University of Bonn, Germany.  
  M. Bennewitz, \mbox{S. Pan}, N. Dengler, and L. Kutsch are additionally with the Lamarr Institute for Machine Learning and Artificial Intelligence and the Center for Robotics, Bonn, Germany. 
  This work has partly been supported by the German Federal Ministry of Research, Technology and Space (BMFTR) under the Robotics Institute Germany (RIG).
  }%
}

\begin{document}
\maketitle
\thispagestyle{empty} 
\pagestyle{empty}

\begin{abstract}
Trajectory optimization is a central component of fast and efficient autonomous racing. 
However practical optimization pipelines remain highly sensitive to initialization and may converge slowly or to suboptimal local solutions when seeded with heuristic trajectories such as the centerline or minimum-curvature paths. 
To address this limitation, we leverage expert driving behavior as a initialization prior and propose a learning-informed initialization strategy based on real-world Formula~1 telemetry.
To this end, we first construct a multi-track Formula~1 trajectory dataset by reconstructing and aligning noisy GPS telemetry to a standardized reference-line representation across 17 tracks.
Building on this, we present a neural network that predicts an expert-like raceline offset directly from local track geometry, without explicitly modeling vehicle dynamics or forces. 
The predicted raceline is then used as an informed seed for a minimum-time optimal control solver.
Experiments on all 17 tracks demonstrate that the learned initialization accelerates solver convergence and significantly reduces runtime compared to traditional geometric baselines, while preserving the final optimized lap time.
\end{abstract}

\section{Introduction}
\label{sec:intro}

Trajectory optimization is a central problem in autonomous racing and general vehicle control~\cite{paden2016survey, betz2022autonomous}. However, solving for globally optimal racing trajectories remains challenging in practice due to the extreme nonlinearity of vehicle dynamics, tire behavior, and track constraints~\cite{pacejka2005tire, rajamani2006vehicle, perantoni2014optimal}.
In particular, many optimization strategies are highly sensitive to initialization, as poor initial trajectories often lead to slow convergence, local minima, or solver failure~\cite{betts2010practical}. 

A common strategy is to initialize optimization using heuristic geometric trajectories, such as the track centerline paths~\cite{heilmeier2020minimum, kapania2016sequential, rosolia2017learning}. 
While computationally cheap, these trajectories are typically far from the optimal racing line, especially in complex corner sequences such as chicanes or closely spaced S-bends, where braking, turn-in, and acceleration phases strongly interact across multiple corners~\cite{perantoni2014optimal}.
As a result, optimization initialized from such baselines may require many iterations to result in competitive solutions or may converge only to a local optimum.

\begin{figure}[t]
  \centering
 \includegraphics[width=0.99\linewidth, trim= 150 50 150 50 , clip]{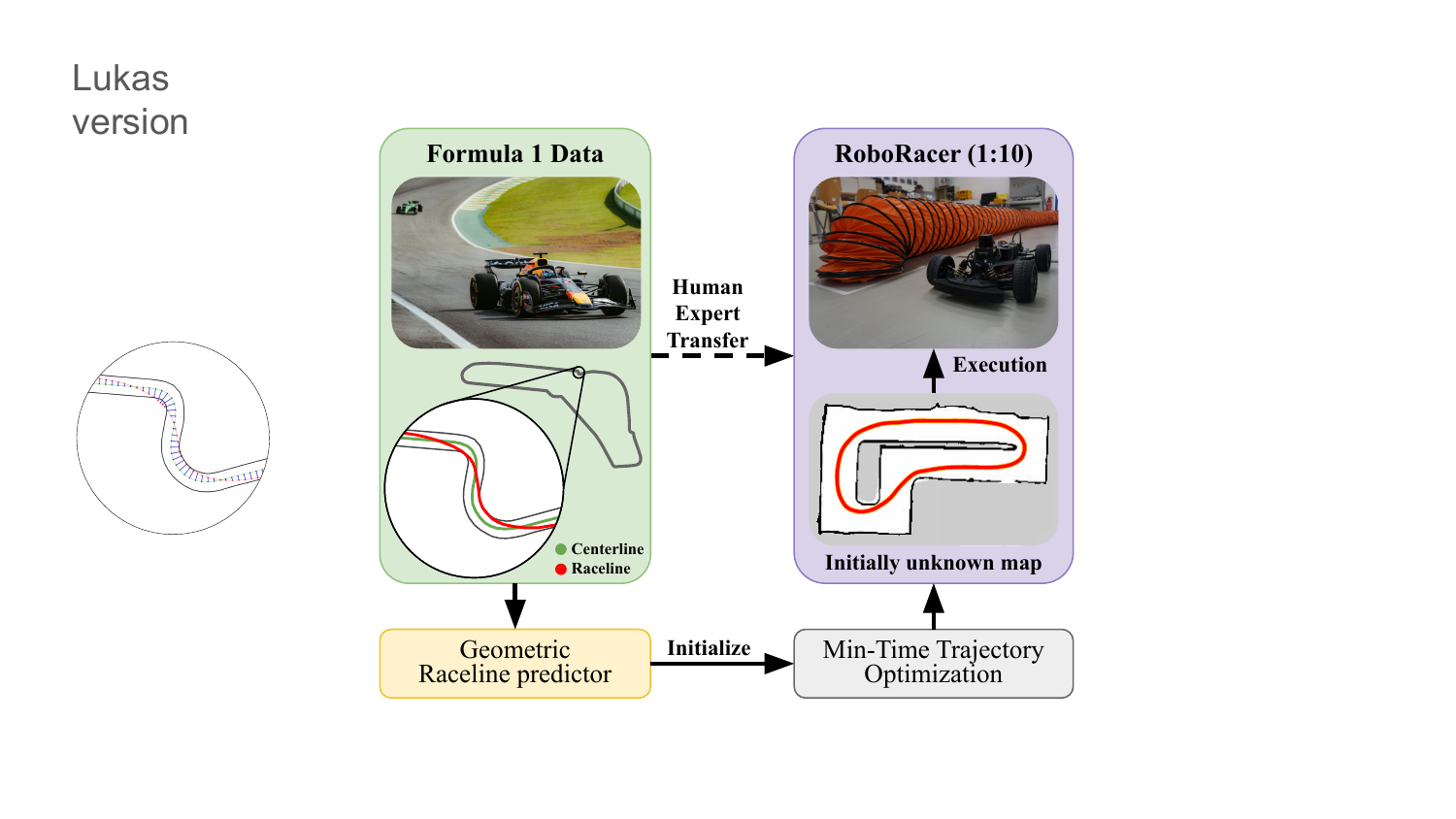}
 \vspace{-15px}
 \caption{Overview of our proposed learning-informed initialization method. 
 First, expert Formula 1 telemetry (left) is used to train a geometric raceline predictor that captures structural properties of near-optimal racing behavior. The predicted raceline then serves as an initialization for a minimum-time trajectory optimization solver. Finally, the resulting optimized trajectory is transferred to a RoboRacer~\cite{charles2025advancing} platform (right), demonstrating robustness to domain shift from full-scale Formula 1 data to a 1:10 autonomous racing system operating on an initially unknown map.}
  \label{fig:motivation}
  \vspace{-10px}
\end{figure}

One source of such informed initializations is human expert driving data. 
Professional drivers operate near the physical limits of the vehicle and track, producing trajectories that reflect implicit trade-offs between corner entry speed, exit speed, curvature, and vehicle stability. 
In particular, \mbox{Formula~1~(F1)} racing lines represent near-optimal solutions for a given vehicle–track combination.
Using these trajectories as initialization can provide a strong prior for optimization, neglecting the need of the solver to discover the global structure of the optimal solution from scratch.

To address the sensitivity of trajectory optimization to initialization quality, we introduce a learning-based raceline prediction approach that leverages real-world F1~telemetry~(see Fig.~\ref{fig:motivation}).
Beyond improving optimization initialization~\cite{pulver2021pilot}, our goal is to enable fast raceline inference on tracks where no expert trajectory is available. 
To this end, we construct a novel multi-track dataset by reconstructing and aligning expert racelines from publicly available race telemetry into a standardized track-centered representation.
Based on this dataset, we propose a network model that predicts a geometric racing line directly from track geometry, without explicitly modeling vehicle dynamics or tire forces.
By learning recurring structural patterns from expert demonstrations, the network provides a data-driven prior that captures realistic driving strategies and trade-offs, and can be used as an informed initialization for trajectory optimization.

Finally, we use the predicted raceline as an informed initialization for a minimum-time optimal control solver~\cite{christ2021time}.
Starting from the learned geometric seed, the optimizer refines the trajectory to satisfy dynamic feasibility and minimize lap time.
We evaluate the proposed initialization strategy against conventional baselines, including the centerline and minimum-curvature trajectories.
Across multiple Formula~1 tracks, our approach improves solver convergence speed while preserving lap time performance.

Our contributions can be summarized as follows:
\begin{itemize}
    \item The introduction of a large-scale Formula 1 trajectory dataset for autonomous racing optimization.
    \item A data-driven initialization strategy for minimum-time racing trajectory optimization, leveraging supervised learning to improve convergence and solution quality.
    \item A systematic evaluation of convergence behavior and lap-time performance, highlighting the role of initialization quality.
\end{itemize}









\section{Related Work}
The Minimum Lap Time Problem (MLTP) aims to compute control inputs that minimize lap time for a given vehicle and track while satisfying track limits and dynamic feasibility~\cite{christ2021time, betz2022autonomous}. 
MLTP is commonly studied within Race Trajectory Optimization (RTO), where methods are typically categorized as free-trajectory or fixed-trajectory approaches depending on whether path and speed are optimized jointly or sequentially~\cite{heilmeier2020minimum, eshof2025computationally}.

Free-trajectory methods formulate MLTP as a minimum-time optimal control problem and solve it via direct transcription into a nonlinear program (NLP), often using interior-point solvers such as IPOPT~\cite{christ2021time, perantoni2014optimal, perantoni2015optimal}. 
While these approaches can produce highly accurate trajectories using detailed vehicle and tire models, they often require computation times on the order of minutes, limiting real-time applicability~\cite{lenzo2020simple, perantoni2015optimal, limebeer2015optimal}. Moreover, NLP-based solvers are highly sensitive to initialization and can converge slowly or to suboptimal local minima when seeded poorly~\cite{kelly2017introduction, betts2010practical}.

In contrast, fixed-trajectory methods reduce computational complexity by decoupling the problem into geometric raceline optimization followed by velocity profile computation under quasi-steady-state assumptions.
Kapania~\etal~\cite{kapania2016sequential} proposed alternating curvature and lap-time optimization, while Heilmeier~\etal~\cite{heilmeier2020minimum} formulated curvature minimization as a quadratic program with improved curvature computation. These methods typically achieve computation times of seconds by computing velocity profiles with forward–backward solvers subject to acceleration limits.
However, comparative studies highlight the resulting trade-off between efficiency and optimality. 
Van den Eshof~\etal~\cite{eshof2025computationally} benchmarked minimum-curvature methods~\cite{heilmeier2020minimum, kapania2016sequential} against joint minimum-time optimization and reported a measurable lap-time deficit for fixed-trajectory approaches, attributing this gap to the missing coupling between path curvature and achievable speed.
However, while these works compare modeling assumptions and lap-time performance, they do not systematically analyze how the choice of geometric initialization influences convergence behavior in minimum-time solvers.

Recent work has explored strategies to overcome the limitations by improving numerical behavior and runtime of classical formulations, including quasi-steady-state reformulations~\cite{christ2021time} and spline-based parameterizations to improve solver stability~\cite{xue2023spline}.
Warm-starting has also been explored in related nonlinear optimization settings to reuse solutions across repeated instances~\cite{banerjee2025deep}, but does not address single-shot optimization on unseen tracks.

Finally, learning-based approaches have been proposed to approximate or complement racing pipelines.
\mbox{Garlick~\etal~\cite{garlick2022real}} proposed networks that directly predict racing trajectories for fast inference, while simulation benchmarks based on human driving facilitate data-driven evaluation~\cite{remonda2024simulation}.
In a related direction, learned model predictive control methods iteratively improve lap performance using previously executed trajectories~\cite{rosolia2017learning}, but typically require repeated laps and online data collection.

In contrast to these methods, our approach retains a physics-based minimum-time optimal control formulation and employs learned racelines solely as geometric initialization priors. 
Rather than replacing the optimization backend or relying on iterative online learning, we aim to improve solver convergence and robustness by injecting structural information derived from expert Formula~1 telemetry into the initialization stage of the optimization process.

\section{Problem Definition}
\label{sec:PD}
The goal of racing trajectory optimization is to compute a dynamically feasible trajectory that minimizes lap time while satisfying track and vehicle constraints.
In this work, we formulate this problem as a continuous Optimal Control Problem~(OCP), defined along the track arc-length~$s$ between two racing waypoints $w_n, w_{n+1}\in W$. 
The optimization variables are the planar racing trajectory, expressed as $x(s)$ and $y(s)$, and the speed profile $v(s)$.
When optimized, the solution must satisfy two main types of constraints: (i) the trajectory must remain within the track boundaries at all times and (ii) the motion must satisfy the vehicle dynamics, which limit feasible accelerations, velocities, and tire forces. 

To evaluate the influence of different initialization strategies on convergence behavior and final lap time performance, we employ a state-of-the-art minimum-time optimal control solver~\cite{christ2021time}. 
Our novel track representation allows the optimizer to enforce boundary constraints directly in the track-aligned frame.

\section{Formula 1 Trajectory Dataset}
To enable data-driven learning of expert racing behavior, we construct a standardized \mbox{Formula~1~(F1)} trajectory dataset by reconstructing and aligning real-world telemetry to track-consistent geometric representations.
In particular, professional motorsport racing data provides near-optimal driving behavior under real-world conditions.
In this context, F1 represents driving at performance limits, making it valuable for learning high-quality racing line geometry. 
However, to the best of our knowledge, no standardized dataset of processed racing trajectories aligned to track geometry currently exists. 
Therefore, we propose a raceline reconstruction pipeline that converts raw telemetry into a track-consistent racing line representation and construct our own dataset, as can be found on github\footnote{\url{https://github.com/HumanoidsBonn/f1-optimization}}.

\subsection{Track Representation}
\label{sec:representation}
\begin{figure}[t]
  \centering
 \includegraphics[width=0.9\linewidth]{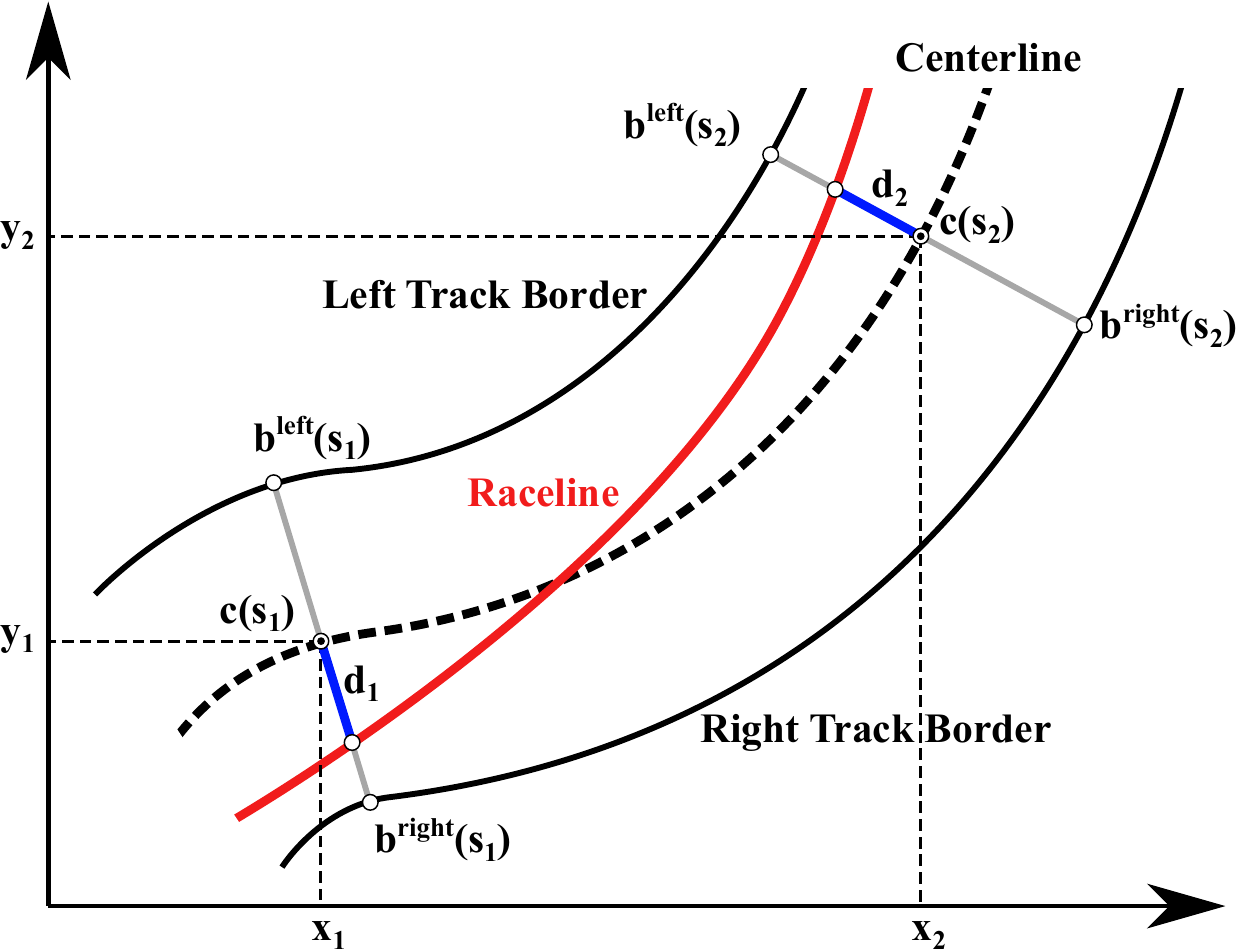}
\caption{The racing track is described by a centerline parameterized by the arc length, with track boundaries and racelines expressed as lateral offsets relative to this reference. 
 }
  \label{fig:track_repr}
  \vspace{-10px}
\end{figure}

To enable a compact and consistent formulation of the~OCP, we use a reference-line-based representation of the track geometry, as shown in \figref{fig:track_repr}.
We model the racing track by using a reference line~$c(s)$ parameterized by its arc-length~$s$. 
The local geometric properties of the track are then described by the curvature $\kappa(s) = \frac{d\theta}{ds}$, where $\theta$ denotes the heading angle of $c$, and the track boundaries $b_{\text{left}}(s)$ and $b_{\text{right}}(s)$, defined as the normal offsets from $c$ to the left and right track bounds respectively.
Although positions within the track can be expressed through global cartesian coordinates, we additionally express each position within the local Frenet-frame~$(s,d)$ relative to the reference line. 
Here, $d$ denotes the lateral offset from the centerline, where a racing line can be compactly expressed as a scalar function $d(s)$.
Note that we standardized all parts of our dataset according to this representation definition.

\subsection{Racetrack and Raceline Extraction}
To reliably obtain data of different F1 race tracks, we used two primary data sources. 
First, we extracted track geometries from the Assetto Corsa racing simulation, a realistic simulator that provides laser-scanned track models and has already been used in other research~\cite{remonda2024simulation}. 
Furthermore, we used the TUM racetrack database\footnote{\url{https://github.com/TUMFTM/racetrack-database}\label{footnote_1}} to obtain additional track geometries. 
We removed tracks that showed a mismatches between the available F1~raceline data and the track geometry.
In total, we extracted geometry data from 17 F1 tracks for our dataset~(see Table~\ref{tab:used_racetracks}). 

\begin{table}[t]
\centering
\begin{tabular}{llll}
\hline
\textbf{Track} & \textbf{Source} & \textbf{Season} & \textbf{Driver} \\
\hline
Baku        & AC  & 2025 & Max Verstappen \\
Barcelona   & AC  & 2025 & Oscar Piastri \\
COTA        & AC  & 2025 & Max Verstappen \\
Hungaroring & AC  & 2025 & Charles Leclerc \\
Imola       & AC  & 2025 & Max Verstappen \\
Interlagos  & AC  & 2025 & Lando Norris \\
Jeddah      & AC  & 2025 & Oscar Piastri \\
Melbourne   & AC  & 2024 & Carlos Sainz \\
Mexico      & TUM & 2025 & Lando Norris \\
Montreal    & TUM & 2025 & George Russell \\
Monza       & AC  & 2025 & Max Verstappen \\
Sakhir      & TUM & 2025 & Oscar Piastri \\
Shanghai    & AC  & 2025 & Oscar Piastri \\
Silverstone & TUM & 2025 & Oscar Piastri \\
Singapore   & AC  & 2025 & George Russell \\
Spa         & TUM & 2025 & Oscar Piastri \\
Spielberg   & AC  & 2025 & Lando Norris \\
\hline
\end{tabular}
\caption{Used racetracks for raceline reconstruction and dataset generation. Track geometries were obtained from the Assetto Corsa racing simulation (AC) and the TUM Racetrack Database (TUM)~\ref{footnote_1}. Melbourne 2024 was used due to race-wide rain conditions in 2025.}
\vspace{-10px}
\label{tab:used_racetracks}
\end{table}

\begin{figure*}[t]
    \centering
    \includegraphics[width=\textwidth]{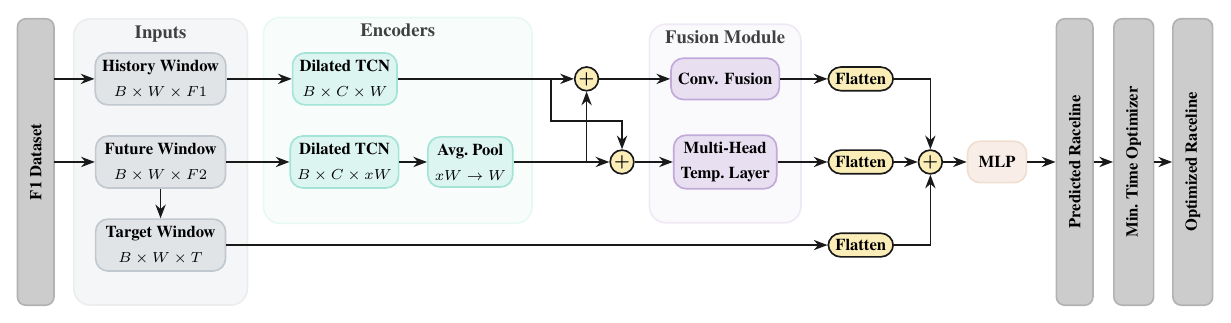}
    \vspace{-20px}
    \caption{Network architecture for raceline prediction. History and future windows are encoded using dilated temporal convolutional networks and fused with the target window via convolutional and multi-head temporal attention layers. The fused features are flattened and processed by an MLP to predict a raceline, which is subsequently used as initialization seed for a minimum-time optimizer.}
    \label{fig:cnn_architecture}
    \vspace{-10px}
\end{figure*} 

To obtain  geometrically consistent F1~data for each of the considered race tracks, we reconstruct racelines from real-world race telemetry and transform them into the track-aligned Frenet frame for trajectory planning.
The telemetry data is obtained using the FastF1 library\footnote{\url{https://github.com/theOehrly/Fast-F1}}, which has been used in prior research for motorsport data analysis and modeling~\cite{eshof2025computationally, sasikumar2025data, mistry2025autof1}, and provides global planar GPS position measurements at approximately~3\,Hz.
In this work, we use telemetry race-day data from race sessions of 2024 and 2025. 
To increase spatial sampling density, we combine telemetry from 15 consecutive laps of a single driver by computing the mean trajectory across all laps.
The driver with a fast time and highest number of valid laps is chosen. 
Table~\ref{tab:used_racetracks} summarizes the selected drivers and years. 

To ensure representative driving behavior, we filter telemetry segments to include only laps that are not affected by rain, safety car phases, and without nearby traffic, i.e., when no other car is directly in front or behind.
Additional preprocessing removes duplicate samples according to their GPS position similarity.
Since the GPS telemetry is provided in global coordinates and may not be aligned with the track reference frame, we perform a global similarity registration between telemetry data and track geometry. 
In particular, let \mbox{$\mathcal{P}=\{p_i\}_{i=1}^N$} denote the telemetry raceline points and \mbox{$\mathcal{C}=\{c_j\}_{j=1}^M$} the track reference geometry, i.e., the centerline. 
We estimate the optimal rotation $R(\theta)$, translation~$t$, and an optional scale $\sigma$ by minimizing a nearest-neighbor alignment loss:
\begin{equation}
\begin{aligned}
\mathcal{L}_{align}
&= \min_{\theta,\,t,\,\sigma}
\frac{1}{N}\sum_{i=1}^N 
\left\| \sigma R(\theta)\,p_i + t - c_{\pi(i)} \right\|_2, \\
\pi(i) 
&= \arg\min_j 
\left\| \sigma R(\theta)\,p_i + t - c_j \right\|_2 .
\end{aligned}
\end{equation}
The optimization is solved using a coarse-to-fine grid search strategy. 
Here, we first perform a coarse search over the rotation angle, followed by local refinement around the best candidate solution. 
Consequently, the transformation yielding the minimum alignment loss is selected. 
After alignment, the raceline is projected onto the corresponding tracks reference line to compute the lateral offset representation. 
The offset between raceline and centerline is computed using cubic spline interpolation~\cite{maeland1988comparison}.

\subsection{Dataset Structure}
Overall, our proposed reconstruction pipeline yields a unified and standardized representation of expert Formula~1 racing trajectories across multiple circuits according to the reference-line formulation described in Sec.~\ref{sec:representation}.
Each track is resampled at a constant spatial resolution of $\Delta s = 2.0\,\text{m}$ along arc length and represented by uniformly spaced waypoints.
For every track, the dataset contains $c(s)$ in global Cartesian coordinates,$\kappa(s)$, and the border offsets relative to $c(s)$. 
In addition, the reconstructed expert raceline is stored as a lateral offset $d_{\text{race}}(s)$, aligned with the same arc-length discretization to ensure that track geometry and expert trajectories are directly comparable across circuits of varying length and complexity. 

\section{Learning-Based Raceline Prediction}
\label{sec:raceline}
Building on the reconstructed F1 trajectory dataset, we learn a data-driven mapping from local track geometry and expert driving history to future raceline geometry.

\subsection{Network Architecture}
The goal of our proposed raceline network is to predict the lateral offset $d_{\text{target}}$ from the track centerline within a given target look-ahead window~(see Fig.~\ref{fig:cnn_architecture}). 
To this end, we us a sliding window approach, i.e., the network processes two centerline-aligned input windows, a history raceline and a future racetrack window.
Both windows contain local track geometry features, including the reference-line curvature $\kappa$ and the left and right track border offsets $b_{\text{left}}$ and $b_{\text{right}}$. In addition, the history window includes the expert raceline offset~$d_{\text{race}}$, providing contextual information about the expected driving style.
 The target window is defined as the first $T$ samples of the future window and represents the prediction horizon of the network.
 Note that the goal of our system is to generate a strong a priori raceline and not to generate an online control mechanism.

As shown in Fig.~\ref{fig:cnn_architecture}, each input window is encoded using a dilated Temporal Convolutional Network (TCN)~\cite{lea2017temporal} composed of two residual 1D convolutional blocks with batch normalization. The use of dilation allows the network to capture both local geometric features and broader contextual structure along the track while maintaining efficient computation.
The encoded history and future features are then combined using two complementary fusion mechanisms. A convolutional fusion module captures local cross-window interactions, while a multi-head temporal attention layer models longer-range dependencies between past driving behavior and upcoming track geometry. 
The fused features are flattened and concatenated with the target window representation, forming a unified feature vector. This representation is processed by a two-layer multilayer perceptron (MLP), which predicts the raceline offset $d(s)$ for the target segment. 

To reconstruct a full-lap raceline, the network predicts the raceline offset for one target segment at a time. After each prediction, the input windows are shifted forward along the centerline of the track and the process is repeated. 
The predicted segments are then concatenated to form the global raceline representation.

\subsection{Training Setup}

During training, the reconstructed F1 racelines are used as ground-truth data. 
The history window contains the ground-truth raceline offset, while the network learns to predict the offset for the upcoming target segment based solely on track geometry and contextual history. 
We split the dataset by track, using 14 tracks for training and 3 tracks for testing cross-track generalization and train the network using a weighted hybrid loss function that combines global trajectory alignment and pointwise spatial accuracy. The loss is defined~as:
\begin{equation}
\mathcal{L}_{train} = 0.5 \cdot \left(1 - \cos(p_{\text{flat}}, g_{\text{flat}})\right)
+ 0.5 \cdot \text{mean}\left( \lVert p - g \rVert_2 \right),
\end{equation}
where $p$ and $g$ denote the predicted and ground-truth raceline offsets, respectively. 
The cosine alignment term encourages global directional consistency of the predicted trajectory sequence, while the Euclidean distance term enforces local geometric accuracy in metric space. The final loss is computed as the mean of both components.


\section{Experimental Evaluation}
To validate the proposed learning-informed initialization strategy, we conduct experimental evaluations across all Formula~1 tracks contained in our dataset. 
Our goal is to quantify how expert data–driven raceline priors influence both geometric prediction quality and downstream trajectory optimization performance. 
In particular, we first assess the geometric quality in comparison to the ground truth data of the predicted trajectories themselves and then investigate how different initialization strategies affect solver convergence speed, robustness, and final lap time performance. 
Furthermore, we perform a hardware validation on a RoboRacer platform~\cite{charles2025advancing} to demonstrate the practical applicability of the proposed approach in a real-world autonomous racing system showing adaptability also to a significant domain shift.

\subsection{Minimum-Time Trajectory Optimization}
\label{sec:optimizer}
To evaluate how different initialization strategies influence convergence behavior, we employ the minimum-time optimal control formulation by Christ~\etal~\cite{christ2021time} as the common optimizer for all experiments.
Longitudinal forces are determined by braking and driving control inputs, while lateral tire forces are modeled using a simplified version~\cite{pacejka2005tire}. 
The model includes vehicle mass, inertia, wheelbase, and tire parameters, which are chosen to approximate a real Formula~1 vehicle. 
For simplicity, a constant tire-road friction coefficient is assumed along the track.
The resulting solution corresponds to a dynamically feasible racing trajectory and velocity profile that minimizes lap time.
We solve the OCP using a nonlinear program (NLP) and optimize it using the IPOPT interior-point solver.
For a detailed mathematical derivation and implementation specifics, refer to~\cite{christ2021time}.
After convergence, the solver outputs a time-optimal racing trajectory together with its corresponding velocity profile. 


\begin{figure*}[t]
  \centering
 \includegraphics[width=0.85\linewidth]{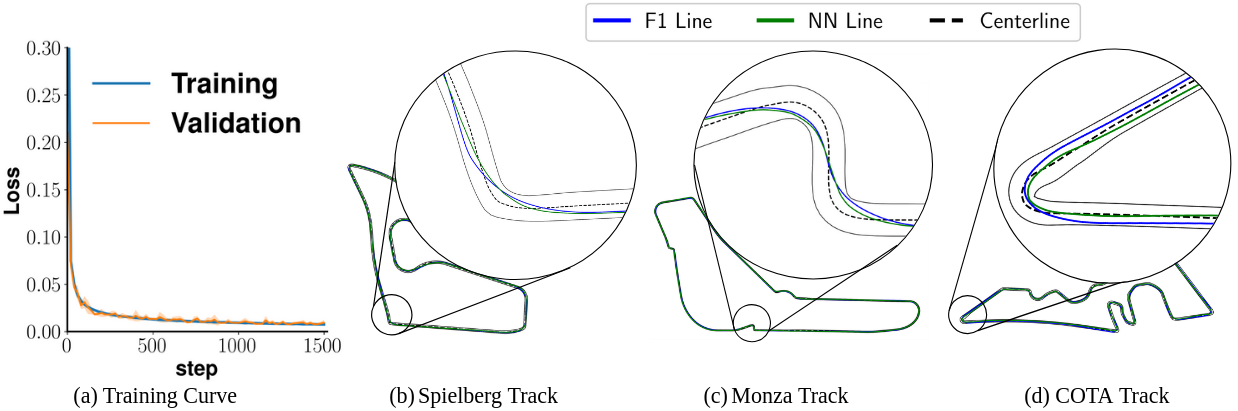}
\caption{Neural network training behavior and qualitative raceline prediction results. 
(a) Training and validation loss curves show stable convergence and similar trends, indicating good generalization without overfitting. 
(b–d) Predicted racelines (green) compared to the reconstructed expert Formula~1 raceline (blue) and the track centerline (black dashed) on representative circuits (Spielberg, Monza, and COTA). 
The predicted trajectories closely follow the expert racing line, particularly in corner entry and exit regions, while clearly deviating from the geometric centerline, demonstrating that the network captures expert driving structure from track geometry alone. 
On COTA the deviation is slightly higher, due to the long straights before the curve where acceleration profiles are difficult to predict.}\label{fig:training_results}
\vspace{-5px}
\end{figure*}

\subsection{Baselines and Metrics}
In total, we compare three different trajectory initialization strategies: (i) the \textbf{centerline (CL)} extracted directly from the track geometry serves as a standard baseline in trajectory optimization due to its simplicity and guaranteed feasibility with respect to track boundaries, (ii) the \textbf{minimum-curvature (MC)} trajectory which is more geometrically refined and generated using the optimization method proposed by Heilmeier et al.~\cite{heilmeier2020minimum}, where curvature minimization is formulated as a quadratic program and yields a geometrically smoother path that more closely resembles typical racing behavior, and  
(iii) our raceline predictor (\textbf{F1-NN}) obtained from the neural network described in Sec.~\ref{sec:raceline}. 
To ensure a fair evaluation of generalization performance, we train the network six times using different 14/3 track-level splits such that each track is excluded from training at least once. For each evaluated track, we therefore use a model that has not been trained on that specific track. As we assume that no expert raceline is available for the tracks during inference we initialize the first window with the centerline and our network predicts the rest of the track.

To assess geometric prediction quality, we compare the predicted racelines against the reconstructed expert trajectory~(\textbf{F1-GT}) using the Root Mean Square Error (\textbf{RMSE}) and Mean Absolute Error (\textbf{MAE}) of the lateral offset $d(s)$ in the Frenet frame. 
Furthermore, to evaluate optimization performance, we measure the number of IPOPT iterations required until convergence, the total solver runtime, and the final optimized lap time after convergence. All initialization strategies are evaluated under identical solver settings to ensure fair and consistent comparison.


\begin{table}[t]
\centering
\begin{tabular}{lcc}
\hline
\textbf{Method} & \textbf{RMSE [m]} & \textbf{MAE [m]} \\
\hline
CL (Centerline) & 3.07 $\pm$ 0.61 & 2.57 $\pm$ 0.55 \\
MC (Min Curve) & 3.12 $\pm$ 0.72 & 2.57 $\pm$ 0.61 \\
F1-NN (Ours)       & \textbf{3.02 $\pm$ 0.69} & \textbf{2.45 $\pm$ 0.63} \\
\hline
\end{tabular}
\caption{Mean geometric deviation with respect to the reconstructed expert raceline (F1-GT) across all 17 tracks of our dataset. RMSE and MAE are computed in the Frenet frame using the lateral offset $d(s)$. Although the geometric improvement of F1-NN over CL is moderate, this small structural offset proves crucial for improving initialization quality in nonlinear minimum-time trajectory optimization. Notably, geometric agreement alone does not fully explain optimization performance, highlighting that initialization quality depends on structural properties beyond simple pointwise deviation.}
\vspace{-10px}
\label{tab:prediction_perform}
\end{table}

\begin{table*}[ht]
\centering
\small
\setlength{\tabcolsep}{4pt}
\resizebox{.8\textwidth}{!}{
\begin{tabular}{lcccc|c}
\hline
\textbf{Init.} 
& \textbf{Iter. $\downarrow$} 
& \textbf{Opt. Time [s] $\downarrow$} 
& \textbf{Gen. Time [s] $\downarrow$} 
& \textbf{Total Runtime [s] $\downarrow$}
& \textbf{Lap Time [s] $\downarrow$} \\
\hline
CL & 521.6 $\pm$ 149.0 & 149.5 $\pm$ 47.5 & -- & 149.5 $\pm$ 47.5 & 85.70 $\pm$ 11.20\\
MC & 483.1 $\pm$ 112.1 & 136.2 $\pm$ 38.7 & 121.4 $\pm$ 50.3 & 257.6 $\pm$ 63.4 & 85.26 $\pm$ 11.17\\
F1-NN (Ours) 
& \textbf{434.5} $\pm$ 103.1$^*$
& \textbf{123.4} $\pm$ 40.5$^*$
& \textbf{0.63} $\pm$ 0.1$^\dagger$
& \textbf{124.0} $\pm$ 40.5$^{*\dagger}$
& \textbf{85.22} $\pm$ 11.21$^*$\\
\hline
F1-GT & 400.1 $\pm$ 67.6 & 112.3 $\pm$ 26.2 & -- & 112.3 $\pm$ 26.2 &85.14 $\pm$ 11.13 \\
\hline
\end{tabular}
}
\caption{Comparison of initialization strategies for minimum-time optimization. 
Reported are the mean and standard deviation of solver iterations, optimized lap time, optimization runtime, initialization generation time, and total runtime across all evaluated tracks. 
Lower values indicate improved efficiency and solution quality. 
CL denotes the centerline, MC the minimum-curvature trajectory, F1-NN the raceline predicted by our neural network, and F1-GT the reconstructed Formula~1 ground-truth trajectory. 
$^{*}$ indicates statistical significance compared to CL ($p < 0.05$), and $^{\dagger}$ indicates statistical significance compared to MC ($p < 0.05$).}
\label{tab:init_results}
\vspace{-10px}
\end{table*}

\subsection{Neural Network Prediction Performance}
First, we evaluate the proposed raceline predictor in terms of (i) training stability, (ii) quantitative geometric accuracy w.r.t.\ the reconstructed expert raceline, and (iii) qualitative behavior on representative circuits.

\subsubsection{\textbf{Training Stability}}
Fig.~\ref{fig:training_results}.a shows smooth and stable convergence of both training and validation loss. The two curves closely follow each other throughout training, indicating well-conditioned optimization and no pronounced overfitting, i.e., the model generalizes to unseen tracks under the track-level split strategy.

\subsubsection{\textbf{Quantitative Geometric Deviation}}
Tab.~\ref{tab:prediction_perform} reports the average RMSE and MAE of the predicted lateral offset~$d(s)$ in the Frenet frame relative to the reconstructed expert raceline (F1-GT). Across all 17 tracks, \textbf{F1-NN} achieves slightly lower error than the geometric baselines. However, these improvements are small and not statistically significant, suggesting that pointwise geometric deviation alone is not sufficient to explain initialization quality for nonlinear minimum-time optimization.

\subsubsection{\textbf{Qualitative Track Behavior}}
Figs.~\ref{fig:training_results}.b--d visualize predicted racelines (green) against the centerline (black dashed) and expert reference (blue) on Spielberg, Monza, and COTA. On Spielberg and Monza, the network reproduces characteristic expert behaviors in the corner entry and exit, clearly deviating from the geometric centerline in a racing-consistent manner. On COTA, residual deviations are larger, due to the long straights before the curve where acceleration profiles are difficult to predict. Overall, these examples illustrate that the model learns circuit-dependent structural patterns that resemble expert driving and provides a racing-oriented geometric prior for downstream optimization.

\subsection{Influence of Initialization Strategies}
To quantify the impact of initialization quality on convergence and final raceline solution quality, we compare different trajectory seeds under identical minimum-time solver settings.
The results averaged for all 17 tracks are summarized in Tab.~\ref{tab:init_results}.

As can be seen, initialization strongly affects convergence efficiency. The centerline baseline requires the most iterations and longest optimization time, while the minimum-curvature (MC) trajectory reduces both metrics by about 7–9\%, confirming the benefit of geometric preprocessing. Our proposed \textbf{F1-NN} initialization further improves convergence, reducing solver iterations by 17\% and optimization time by 26\,s compared to the centerline. This indicates that the learned raceline places the optimizer significantly closer to the optimal solution compared to the centerline.

However, initialization cost must also be considered. 
The MC trajectory requires substantial preprocessing, resulting in the highest total overall runtime.
In contrast, \textbf{F1-NN} generates racelines in only 0.63\,s , yielding the lowest overall runtime, nearly halving total computation compared to MC and reducing runtime by 17\% compared to the centerline baseline. Finally, the small remaining gap to the \textbf{F1-GT} initialization confirms that the network successfully transfers expert trajectory structure into improved optimization performance. 
Overall, these results highlight that learning-based initialization substantially improves solver efficiency while maintaining or slightly improving final trajectory quality, achieving near-expert performance at negligible computational cost.

\subsection{Hardware Validation}
We conducted Real-world experiments on a self-built RoboRacer 1:10 autonomous racing platform~\cite{charles2025advancing, baumann2024forzaeth} to validate practical feasibility of our approach and demonstrate that the proposed \textbf{F1-NN} initialization transfers from simulation to a real system, despite significant domain shift.
Therefore, we constructed a dedicated test track with an L-shaped layout, including straights and both left- and right-hand corners to cover varying curvature conditions. 

To generate global racelines, we use the same minimum-time optimal control solver described in Sec.~\ref{sec:optimizer}, parameterized with RoboRacer-specific vehicle dynamics. The proposed \textbf{F1-NN} model, trained exclusively on full-scale Formula~1 data, predicts an initialization raceline for the experimental track, representing a substantial domain shift to a previously unseen, downscaled environment. The vehicle tracks the optimized trajectory using a predictive pure pursuit controller operating on the minimum-time trajectory and corresponding speed profile.

In this experiment, we compare centerline initialization~(\textbf{CL}) against the proposed \textbf{F1-NN} initialization by executing the resulting optimized racelines and velocity profiles on the vehicle. Each configuration is evaluated over 10~autonomous laps under identical environmental conditions. Feasibility is assessed by counting crash-free runs, while performance metrics are derived from recorded lap times and tracking data. In addition, we report the lateral tracking error, defined as the perpendicular distance between the planned raceline and the executed vehicle trajectory, which reflects how well the controller can follow the optimized path.

\begin{figure}[t]
  \centering
 \includegraphics[width=0.9\linewidth]{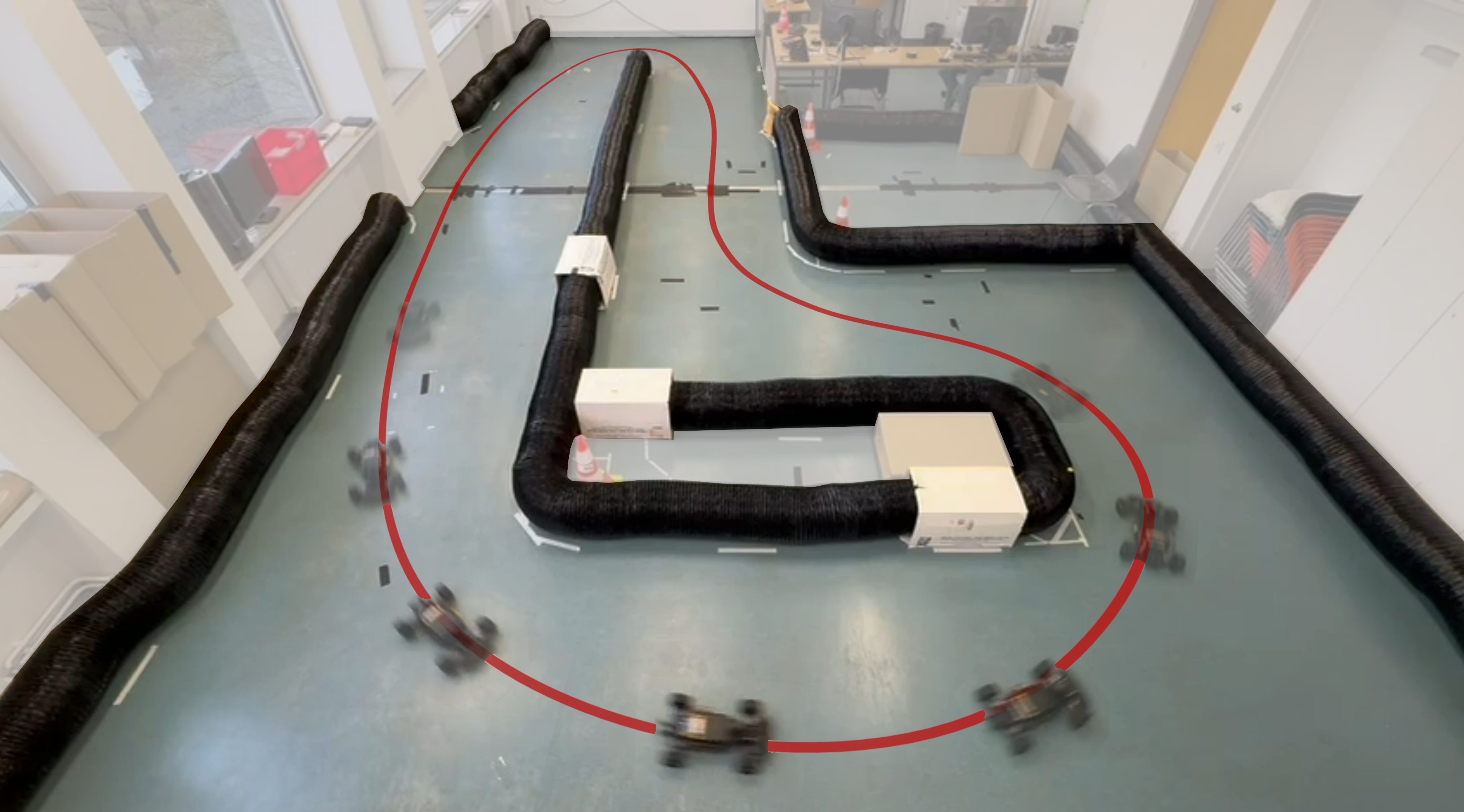}
\caption{Qualitative example of the optimized raceline on the RoboRacer test track. The F1-NN-initialized trajectory (red) follows smooth cornering arcs resulting in high speed and trackability.} 
\vspace{-10px}
\label{fig:real_race}
\end{figure}

The quantitative results in Tab.~\ref{tab:experiment_summary} show that \textbf{F1-NN} initialization achieves significantly faster lap times while also reducing lateral tracking error compared to the centerline baseline. The reduced lateral error indicates that the resulting trajectories are not only faster but also easier for the controller to track reliably, suggesting improved geometric smoothness and dynamic feasibility. Furthermore, the higher average speed confirms that the optimizer can exploit more aggressive yet executable trajectories when initialized with the learned raceline prior. Fig.~\ref{fig:real_race} shows a qualitative example of the optimized trajectory on the real test track, illustrating how the \textbf{F1-NN} initialization produces a smoother and more corner-efficient racing line.

\begin{table}[t]
\centering
\small
\setlength{\tabcolsep}{4pt}
\resizebox{\columnwidth}{!}{
\begin{tabular}{l|c|c|c}
\hline
\textbf{Init.} & \textbf{Lap Time [s]} & \textbf{Speed [m/s]} & \textbf{Lat. Err [m]} \\
\hline
CL & 7.090 $\pm$ 0.061& 3.504 $\pm$ 1.010  & 0.165 $\pm$ 0.155  \\
F1-NN (Ours) & 6.640 $\pm$ 0.069$^*$ & 3.672 $\pm$ 1.014$^*$ & 0.109 $\pm$ 0.075$^*$ \\
\hline
\end{tabular}
}
\caption{Real-world RoboRacer evaluation comparing centerline~(CL) and our proposed F1-NN initialization. Reported are mean $\pm$ standard deviation across 10 autonomous laps. \textbf{Lat. Err} denotes the lateral tracking error between the planned raceline and the executed vehicle trajectory. $^{*}$ indicates statistically significant improvement over CL ($p < 0.01$).}
\label{tab:experiment_summary}
\vspace{-8px}
\end{table}

\section{Conclusion}
\label{sec:conclusion}
In this paper, we presented a learning-informed initialization strategy for minimum-time trajectory optimization in autonomous racing. 
Motivated by the sensitivity of nonlinear optimal control solvers to the initial guess, we use expert Formula~1 driving behavior as an optimization prior.
To this end, we reconstruct real-world Formula~1 telemetry across 17~tracks and train a windowed CNN to predict expert-like raceline geometry from local track features.
Experiments across all tracks show that the learned initialization improves solver efficiency under identical settings. Compared to centerline and minimum-curvature seeds, F1-NN reduces iterations and optimization time while approaching the performance of reconstructed expert-raceline initialization. 
Real-world experiments on a 1:10 RoboRacer platform further confirm transfer under domain shift, yielding faster lap times and lower tracking error than a centerline-based baseline.

As future work, we aim to extend the geometry-conditioned raceline prior with additional context, such as tire state, grip conditions, vehicle setup, weather, and race strategy. This could enable more tactical raceline generation, e.g., for overtaking or tire management. Broader datasets covering multiple drivers, vehicle configurations, and track conditions would further help distinguish universal racing patterns from driver- or vehicle-specific behavior.

\bibliographystyle{IEEEtran}

\bibliography{bibliography}

\end{document}